\documentclass[10pt,twocolumn,letterpaper]{article}

\usepackage{iccv}

\usepackage{breakurl}
\usepackage[utf8]{inputenc} 
\usepackage[T1]{fontenc}    
\usepackage{xurl}        
\usepackage{booktabs, multirow}
\usepackage{amsfonts}       
\usepackage{nicefrac}       
\usepackage{microtype}     
\usepackage{xcolor}        
\usepackage[authoryear]{natbib}
\usepackage{algorithm,algpseudocode}
\usepackage{amsmath}
\usepackage{graphicx}
\usepackage{amsthm}
\theoremstyle{definition}

\usepackage{caption}
\usepackage{subcaption}
\usepackage[breaklinks=true,bookmarks=false]{hyperref}

\iccvfinalcopy 

\title{DeepfakeArt Challenge: A Benchmark Dataset for Generative AI Art Forgery and Data Poisoning Detection}

\author{%
   Hossein Aboutalebi$^{1 *}$, Dayou Mao$^{1 *}$, Rongqi Fan
, Carol Xu, Chris He$^{1}$, and Alexander Wong$^{1,2,3}$ \\
  $^1$ Vision and Image Processing Research Group, University of Waterloo\\
  $2$ Waterloo Artificial Intelligence Institute, Waterloo, Canada\\
  $3$ DarwinAI Corp., Waterloo, Canada\\
  $*$ Equal Contribution\\
}

\begin{document}

\maketitle
\begin{abstract}
\vspace{-0.1in}
The tremendous recent advances in generative artificial intelligence techniques have led to significant successes and promise in a wide range of different applications ranging from conversational agents and textual content generation to voice and visual synthesis. Amid the rise in generative AI and its increasing widespread adoption, there has been significant growing concern over the use of generative AI for malicious purposes. In the realm of visual content synthesis using generative AI, key areas of significant concern has been image forgery (e.g., generation of images containing or derived from copyright content), and data poisoning (i.e., generation of adversarially contaminated images). Motivated to address these key concerns to encourage responsible generative AI, we introduce the DeepfakeArt Challenge, a large-scale challenge benchmark dataset designed specifically to aid in the building of machine learning algorithms for generative AI art forgery and data poisoning detection. Comprising of over 32,000 records across a variety of generative forgery and data poisoning techniques, each entry consists of a pair of images that are either forgeries / adversarially contaminated or not. Each of the generated images in the DeepfakeArt Challenge benchmark dataset \footnote{The link to the dataset: http://anon\_for\_review.com} has been quality checked in a comprehensive manner. 

\end{abstract}
\vspace{-0.2in}
\section{Introduction}
\vspace{-0.1in}
\begin{figure}[!htb]
\vspace{-0.08in}
    \begin{center}
    \includegraphics[width=23em]{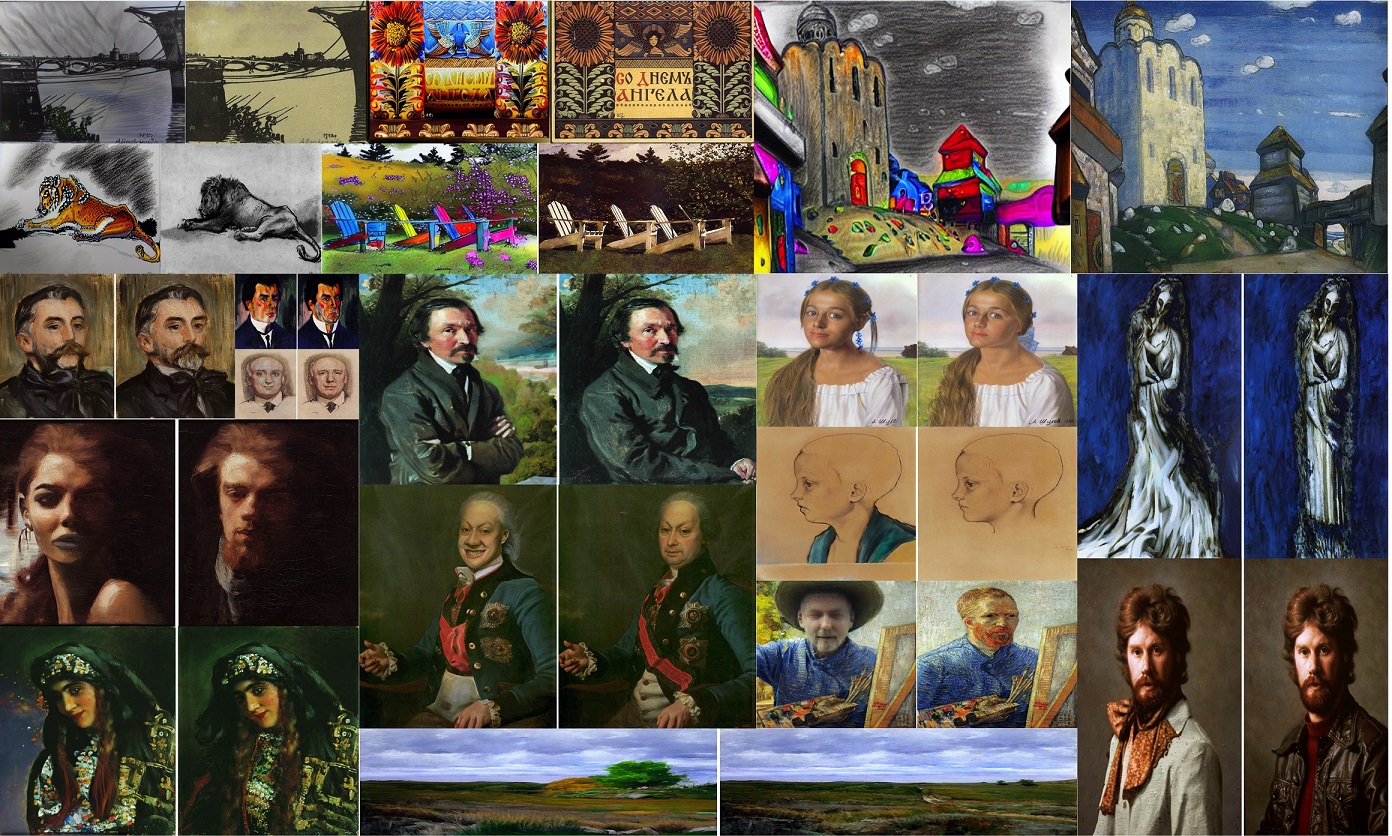}
    \caption{Example images from the proposed DeepfakeArt Challenge dataset.}
    \label{fig:all_class_act}
    \end{center}
    \vspace{-0.4in}
\end{figure}

The significant advances in the field of generative artificial intelligence (AI) made in recent years have resulted in an enormous interest and excitement in both the research community and the general public around the potential to produce very high quality synthetic content.  For example, in the field of natural language processing (NLP), the introduction of High profile endeavors in the Natural Language Processing (NLP) realm, the introduction of foundational large language models (LLMs) such as GPT-4~\cite{openai2023gpt} and LLaMA~\cite{touvron2023llama}, trained on large language datasets, has led to demonstrated ability to generate complex synthetic textual content that highly resemble that produced by humans.  Furthermore, such LLMs have facilitated for the development of conversational AI models such as ChatGPT~\cite{liu2023summary} and Vicuna~\cite{chiang2023vicuna}, which can now generate very detailed and structured answers and carry longer conversations with human users in such a convincing way that was shown, in some studies, to be preferred over that of a human expert~\cite{ayers2023comparing}.

Equally impressive are advancements in generative AI for generating high quality synthetic visual content~\cite{ramesh2021zero,blattmann2023align, rombach2022high}.  For example, the recent introduction and advances in diffusion models \cite{ramesh2021zero,oppenlaender2022creativity}, trained on large image datasets such as LAION \cite{schuhmann2021laion}, has led to the ability to generate highly complex and intricate imagery that rival human artists in art competitions \cite{roose2022ai} and photography competitions \cite{AllisonUnifiedDS}. 

Amid the rise in generative AI due to its tremendous potential and its current trajectory towards widespread adoption, there has been tremendous growing concern over its intentional or unintentional use for malicious purposes that can have a negative impact on society.  For example, in the realm of visual content synthesis using generative AI, there is a growing concern over image forgery and copyright infringement \cite{Will.2022, Ellen.2023, James.2023}.  More specifically, large generative models for image synthesis are typically trained on large image datasets such as LAION \cite{schuhmann2022laion}, which contain billions of images with many of these being copyright-protected images.  Furthermore, given that such models also accept language prompts, they are also trained on textual datasets often comprising of internet corpora, which could contain proprietary or copyright-protected text content. Given the sheer size of these datasets, it is nearly impossible to manually inspect all content within these datasets, leaving the origins and intellectual property rights of the data sources largely ambiguous \cite{somepalli2022diffusion}. A critical concern arising from this is the tendency of large generative models to memorize data, as highlighted in various studies \cite{carlini2022quantifying, biderman2023emergent}. This presents significant legal considerations for companies deploying these models, who must navigate these issues carefully to avoid infringement.  Another growing concern in the realm of visual content synthesis is the rise of adversarial data poisoning intended to fool AI-based systems for decision-making and decision support.  For example, there has been significant recent explorations on the impact of data poisoning in AI-powered medical systems, where the intentional embedding of generated adversarial noise into medical images that are imperceptible to the human eye can lead to false diagnosis that can have lethal consequences for patients as well as other malicious acts such as insurance fraud \cite{ghaffari2022adversarial,finlayson2019adversarial, zhou2021machine}.  As such, means to tackle these concerns are critical for widespread adoption in a responsible manner.
\vspace{-0.04in}
Motivated to address these key concerns to encourage responsible generative AI for visual content synthesis, we introduce the DeepfakeArt Challenge, a large-scale challenge benchmark dataset designed specifically to aid in the building of machine learning algorithms for generative AI art forgery and data poisoning detection. Figure \ref{fig:all_class_act} shows the snapshot examples of our dataset. To the best of the authors' knowledge, there are currently no available datasets designed specifically for the purpose of exploring the detection of copyright infringement for generative AI in the domain of art.  As such, this study represents a first attempt at producing a dataset for supporting researchers in this field in the development of machine learning algorithms for identifying copyright infringement by generative AI models. The development of such detection algorithms could significantly assist developers of generative AI systems in mitigating potential legal repercussions associated with their systems, as well as enable content creators and content curators to better identify copyright infringement by generative AI efficiently and systematically.
\vspace{-0.05in}
\section{Problem Definition}
\vspace{-0.05in}
The two core focuses of the proposed DeepfakeArt Challenge dataset center around art forgery and data poisoning, and as such it is important to establish the problem definition for each of these two core focuses.

\textbf{Notations}. In the following, we denote by $\mathbb{R}^{H \times W \times C}$ an image of height $H$ and width $W$ with color channels $C$. In addition, $[n]$ will denote the set $\{1, ..., n\}$, for any $n \in \mathbb{N}$. When given a region $\Omega \subseteq [H] \times [W] \times [C]$ and an image $I \in \mathbb{R}^{H \times W \times C}$, $I_{\Omega} \in \mathbb{R}^{H \times W \times C}$ will denote the tensor obtained from $I$ by keeping the entries within the region $\Omega$ and suppressing the entries outside to zero.

\subsection{Art Forgery}\vspace{-0.05in}
In the case of art forgery, the goal of the proposed DeepfakeArt Challenge dataset is about the identification of copyright infringement, and as such it is important to set the context along with a formal definition.  Specifically, generative AI models typically utilize vast amounts of training data to achieve strong content generation performance.  However, many of the large-scale datasets used for training such as the LAION datasets \cite{schuhmann2021laion} contain significant amounts of copyright-protected content. Hence, generative AI models trained on such datasets can produce synthetic content that contains or highly resembles copyright-protected content \cite{somepalli2022diffusion}. 

In judicial proceedings, the "substantial similarity" test is employed to ascertain if a party has violated another party's copyright \cite{asay2022empirical}. However, determining substantial similarity remains a subject of ongoing debate due to its inherently open-ended nature. While there's no precise formula, courts do follow certain criteria to arrive at a decision. In particular, the substantial similarity doctrine in copyright law necessitates that a court confirm the defendant's replication is both quantitatively and qualitatively similar enough to the original to warrant classification as infringement \cite{balganesh2014judging}. As discussed in~\cite{balganesh2014judging}, the jury evaluates the evidence based on several factors, including the creativity exhibited, the work's creation process, its intended purpose, the defendant's creative contributions and conduct, and the market impacts of the defendant's copying.

Building on the principles inherent in the "substantial similarity test" used in legal proceedings, let us establish a pragmatic definition of copyright infringement as it applies to generative AI and as applied in the creation of the proposed DeepfakeArt Challenge dataset. This definition is rooted in the exploration and consideration of two foundational conditions.

\textbf{Definition: Copyright Infringement in a Generative AI Model} — Consider a generative model $M: \mathbb{R}^{H \times W \times C} \times \mathcal{Q} \to \mathbb{R}^{H' \times W' \times C'}$ where $\mathcal{Q}$ could represent the set of any additional input to the generative model.
Assume the model $M$ has been trained on some dataset $\mathcal{D} \subset \mathbb{R}^{H \times W \times C}$, within which a subset $\hat{\mathcal{D}} \subseteq \mathcal{D}$ is the set of copyright-protected examples. Given some norm $\|\cdot\|: \mathbb{R}^{H' \times W' \times C'} \to [0, +\infty)$, some threshold $\delta > 0$, and some monotonically non-decreasing function $f: \mathbb{N} \cup \{0\} \to [0, +\infty)$, we classify a generated output $y \in \mathbb{R}^{H' \times W' \times C'}$ by our model $M$ as infringing on copyright, with respect to $\|\cdot\|$, $\delta$, and $f$, if and only if there exists some $\hat{x} \in \hat{\mathcal{D}} \subset \mathbb{R}^{H \times W \times C}$ such that any one of (or both of) the following conditions hold:
\begin{itemize}
    \item There exist some region $\Omega \subseteq [H'] \times [W'] \times [C']$ of a significant size and some transformation $T: \mathbb{R}^{H \times W \times C} \to \mathbb{R}^{H' \times W' \times C'}$, s.t.
    \begin{equation} \label{inpaint_eq}
        \|y_\Omega - (T(\hat{x}))_\Omega\| < 
f(|\Omega|) \cdot \delta.
    \end{equation}

    \item There exist some edge detection model $E: \mathbb{R}^{H' \times W' \times C'} \to \{0, 1\}^{H' \times W' \times C'}$, some region $\Omega \subseteq [H'] \times [W'] \times [C']$ of a significant size, and some transformation $T: \mathbb{R}^{H \times W \times C} \to \mathbb{R}^{H' \times W' \times C'}$, s.t.
    \begin{equation} \label{style_eq}
        \|(E(y))_\Omega-(E(T(\hat{x})))_\Omega\| < f(|\Omega|) \cdot \delta.
    \end{equation}
\end{itemize}
If either condition (\ref{inpaint_eq}) or (\ref{style_eq}) is met, it is concluded that the model $M$ infringes upon the copyright protection for $\hat{x}$, by the intensity of $f(|\Omega|)/\delta$ $\square$

\begin{table*}[t]
\centering
\vspace{-0.05in}
\caption{Performance of similarity detection for different models on DeepFake Art dataset}
\vspace{-0.05in}
\label{tab:sample_table}
\begin{tabular}{lccccc}
\toprule
Model & Accuracy & Precision & Recall & F1 & MAE \\
\midrule
MultiGrain \cite{berman2019multigrain} & 81.20 \% & \textbf{97.68}\% & 63.90 \% & 0.77 & 0.42 \\
DINO-v1 XCiT-S/16 \cite{caron2021dinov1} & 80.84 \% & 97.65 \% & 63.18 \% & 0.76 & 0.32 \\
DINO-v1 XCiT-S/8 \cite{caron2021dinov1}  & 78.75 \% & 97.05 \% & 59.28 \% & 0.73 & 0.32 \\
DINO-v1 ViT-S/8 \cite{caron2021dinov1} & 81.71 \% & 96.74 \% & 65.61 \% & 0.78 & 0.34 \\
DINO-v1 ResNet-50 \cite{caron2021dinov1} & 81.11 \% & 96.22 \% & 64.78 \% & 0.77 & 0.38 \\
DINO-v2 ViT-S/14 \cite{oquab2023dinov2} & 81.12 \% & 96.88 \% & 64.29 \% & 0.77 & 0.31 \\
DINO-v2 ViT-B/14 \cite{oquab2023dinov2} & 81.88 \% & 95.65 \% & 66.78 \% & 0.79 & 0.30 \\
DINO-v2 ViT-L/14 \cite{oquab2023dinov2} & \textbf{82.75} \% & 95.09 \% & \textbf{69.04} \% & \textbf{0.80} & \textbf{0.29} \\
DINO-v2 ViT-g/14 \cite{oquab2023dinov2} & 81.20 \% & 95.05 \% & 65.81 \% & 0.78 & \textbf{0.29} \\
SwinTransformer \cite{liu2021swin} & 78.02 \% & 96.17 \% & 58.34 \% & 0.72 & 0.35 \\
\bottomrule
\end{tabular}
\vspace{-0.05in}
\label{results}
\end{table*}

We make a few remarks on our definition above. 1) The region $\Omega$ is a set of indices of the form $(i, j, k)$. Putting it in subscript suppresses the entries outside the region to zero, so that they will not contribute to the norm. 2) The transformation $T$ can only be composed of geometric transformations including resizing, zooming, translation, rotation, flipping, etc. 3) The smaller the threshold $\delta$ and the larger the region $\Omega$, the higher the infringement intensity $f(|\Omega|) / \delta$. 4) The function $f$ is used to make the threshold $\delta$ be adaptive to the size of the region. Simple forms could take constant ($f(n) := 1$), linear ($f(n) := n$), or square root ($f(n) := \sqrt{n}$).

The basis for the design and protection against copyright infringement is established by conditions (\ref{inpaint_eq}) and (\ref{style_eq}). In this context, the strategies we have implemented in the formation of datasets for inpainting, cutmix, and adversarial data poisoning comply with condition (\ref{inpaint_eq}). In addition, the style transfer technique we have utilized adheres to condition (\ref{style_eq}).

\textbf{Definition: Adversarial Data Poisoning} — For a given prediction model $M: \mathbb{R}^{H \times W \times C} \to \mathcal{Y}$ and input $x \in \mathbb{R}^{H \times W \times C}$, we call $x' \in \mathbb{R}^{H \times W \times C}$ an adversarial input, with respect to some norm $\|\cdot\|: \mathbb{R}^{H \times W \times C} \to [0, +\infty)$ and tolerance $\epsilon > 0$, if and only if both of the following conditions hold:
\begin{equation} \label{adv_cond_1}
    \|x - x'\| < \epsilon \\
\end{equation}
\begin{equation} \label{adv_cond_2}
    M(x) \neq M(x')
\end{equation}

If conditions (\ref{adv_cond_1}) and (\ref{adv_cond_2}) are both satisfied, we call $x'$ the adversarially generated version of $x$ for $M$ \cite{goodfellow2014explaining}.

\vspace{-0.05in}
\section{Methodology}
\label{sec:methodology}
\vspace{-0.05in}
Based on the aforementioned definitions of copyright infringement and adversarial data poisoning in the context of generative AI, the proposed DeepfakeArt Challenge benchmark dataset consists of more than 32,000 records generated using the following generative forgery and data poisoning methods: 1) Inpainting, 2) Style Transfer, 3) Adversarial data poisoning, and 4) Cutmix.  These were generated by modifying source images from the WikiArt dataset \cite{saleh2015large} to produce forgery images. This transformation process is based on four main principles, each representing a different form of copyright violation: 1) The usage of partial image data from the source, with the remaining portion inpainted. 2) The preservation of original image edge strokes, but with a modification in the painting style. 3) The introduction of minor noise into the image. 4) The replication, either partially or entirely, of another image.

\begin{figure}[!htb]
    \begin{center}
    \includegraphics[width=12em]{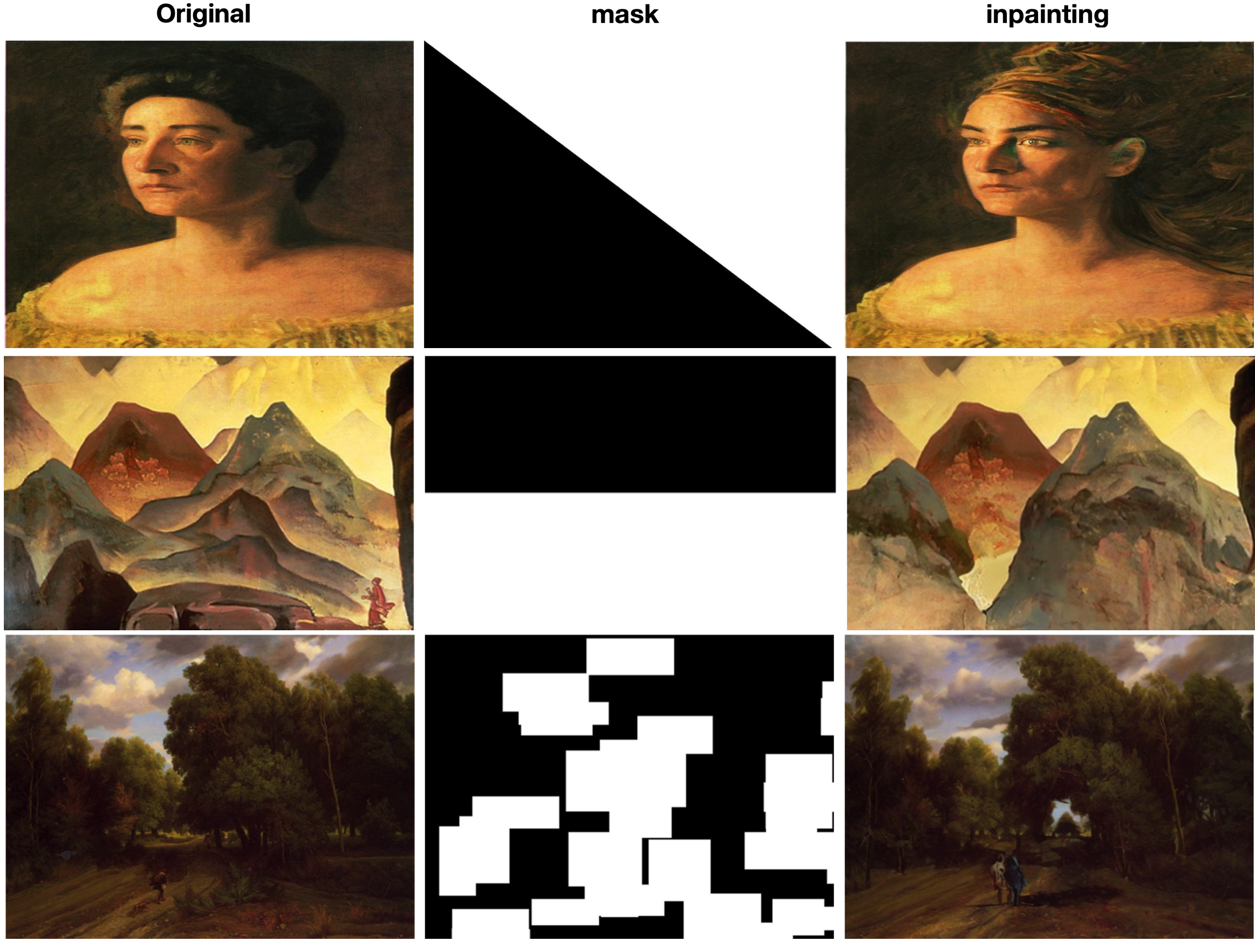}~\includegraphics[width=12em]{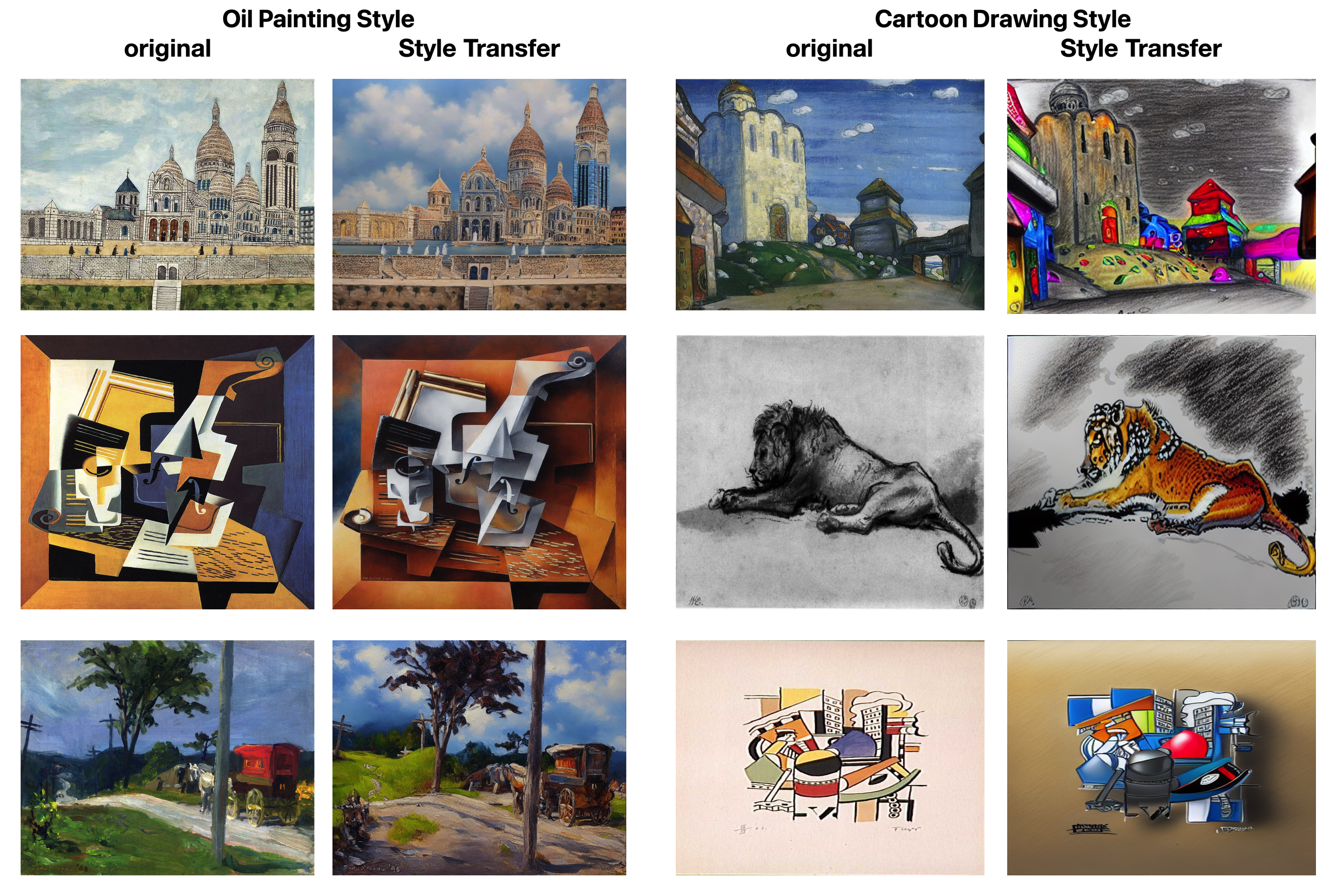}
    \caption{(left) Examples of various masks used for generating forgery pairs in inpainting category. (right) Example generated original-forgery image pairs in the style transfer category.}
    \label{fig:inpainting_fig}
    \end{center}
    \vspace{-0.3in}
\end{figure}

To demonstrate these types of copyright violations, we have established four categories within our dataset:

\begin{itemize}
\item \textbf{Inpainting}: Using Stable Diffusion 2~\cite{rombach2022high}, we inpainted parts of the source image. In each instance, a source image
was randomly chosen from the WikiArt dataset and then inpainted to create the forgery images.  The prompt used for generating the inpainting image was: "generate a painting compatible with the
rest of the image". 

Three different masking techniques were used: 1- Side masking 2- Diagonal masking 3- Random masking. Figure \ref{fig:inpainting_fig}(left) presents examples of the source, mask, and the resulting inpainted images. \vspace{-0.05in}
\item \textbf{Style Transfer}: We employed ControlNet~\cite{zhang2023adding} to alter the style of the source image, utilizing Canny edge detection.
To guide the generation of style-transferred images, we utilized four distinct prompts, each corresponding to a different style: 1) "a high-quality, detailed, realistic image", 2) "a high-quality, detailed, cartoon-style drawing", 3) "a high-quality, detailed, oil painting", and 4) "a high-quality, detailed, pencil drawing".

Figure \ref{fig:inpainting_fig}(right) shows examples of transformed source image to get altered style transfer image.
\vspace{-0.05in}
\item \textbf{Adversarial Data Poisoning}: To introduce noise into the source image, we deployed three adversarial attacks: FGSM~\cite{goodfellow2014explaining}, APGD~\cite{croce2020reliable}, and PGD~\cite{madry2017towards}.\vspace{-0.05in}
\item \textbf{Cutmix}: Taking inspiration from the work of~\cite{somepalli2022diffusion}, we replicated portions of the source image.\vspace{-0.05in}
\end{itemize}

\subsection{Data distribution}
\label{sec:datadistribution}
\begin{figure}[!htb]
     \vspace{-0.01in}
     \begin{center}
    \includegraphics[width=23em]{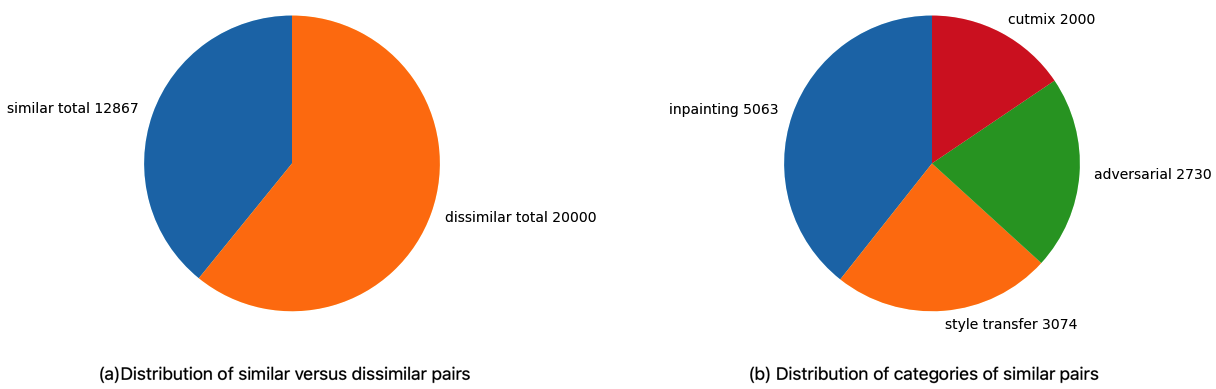}
    \caption{Distribution of data}
        \label{fig:overall_dist}
     \end{center}
     \vspace{-0.3in}

\end{figure}

\vspace{-0.05in}
The DeepfakeArt Challenge benchmark dataset encompasses over 32,000 records, incorporating a wide spectrum of generative forgery and data poisoning techniques. Each record is represented by a pair of images, which could be either forgeries, adversarially compromised, or not. Fig. \ref{fig:overall_dist} (a) depicts the overall distribution of data, differentiating between forgery/adversarially contaminated records and untainted ones. The dispersion of data across various generative forgery and data poisoning techniques is demonstrated in Fig. \ref{fig:overall_dist} (b). Notably, as presented in \ref{fig:overall_dist} (a), the dataset contains almost double the number of dissimilar pairs compared to similar pairs, making the identification of similar pairs substantially more challenging given that two-thirds of the dataset comprises dissimilar pairs.

\vspace{-0.1in}
\section{Experiments and Discussion}
\vspace{-0.05in}
In Table \ref{results}, we present the performance of various models evaluated on the DeepfakeArt dataset. Our task was a classification challenge, wherein the objective was to distinguish between 'similar pairs' (pairs where one image potentially infringes copyright) and 'dissimilar pairs' (pairs of unrelated images).

\noindent\textbf{Evaluation}.
In this study, we evaluated several prominent models: MultiGrain \cite{berman2019multigrain}, DINO-v1 \cite{caron2021dinov1}, DINO-v2 \cite{oquab2023dinov2}, and SwinTransformer \cite{liu2021swin}. We utilized the embeddings generated by these models, calculating the dot product between each pair. Subsequently, we applied cross-validation to pinpoint the optimal threshold to distinguish between similar and dissimilar pairs. Utilizing this threshold, we quantized the calculated cosine similarities into binary outcomes, facilitating the analysis of various metrics, such as accuracy, precision, and recall, all of which are documented in Table \ref{results}.

\noindent\textbf{Discussion}.
As delineated in Table \ref{results}, the DINO-v2 ViT-L/14 model notably outshines the others in overall performance, albeit with a slight drawback in the precision metric, where the MultiGrain model emerges as the frontrunner. Despite showcasing considerable precision, the models generally falter in terms of recall. This indicates a heightened rate of false negatives across all models, a trend which holds significant implications in the context of copyright infringement detection. Particularly, a higher frequency of false negatives could potentially escalate the risk of litigations concerning copyright infringements in generative AI models.

In conclusion, this research articulates a nuanced definition of copyright infringement and introduces a synthetic dataset designed to emulate real-world scenarios where such infringements may occur. The experimental results underscore the current models' tendency to incur high false negative rates when applied to the dataset, thereby highlighting avenues for the development of more robust and efficient detection tools to identify and mitigate copyright infringements.

\sloppy
\bibliographystyle{plainnat}
\bibliography{neurips_2023}

\end{document}